\documentclass[sigconf]{acmart}
\usepackage{multirow}
\usepackage{tabularx}
\usepackage{makecell}
\usepackage{float}
\usepackage{pifont} % put this in the preamble
\AtBeginDocument{%
  }

\setcopyright{acmlicensed}
\copyrightyear{2025}
\acmYear{2025}
\acmDOI{XXXXXXX.XXXXXXX}
\acmConference[CODS '25]{Make sure to enter the correct
  conference title from your rights confirmation email}{Dec 17--20,
  2025}{Pune, India}
\acmBooktitle{Proceedings of the 12th ACM IKDD Conference on Data Science}
\acmISBN{978-1-4503-XXXX-X/2018/06}

\begin{document}

\title{HistoLens: An Interactive XAI Toolkit for Verifying and Mitigating Flaws in Vision-Language Models for Histopathology}


\author{Vissapragada Sandeep}
\affiliation{%
  \institution{Indian Institute of Technology}
  \city{Bhilai}
  \state{Chhattisgarh}
  \country{India}
}
\email{vissapragada@iitbhilai.ac.in}

\author{Vikrant Sahu}
\affiliation{%
  \institution{Indian Institute of Technology}
  \city{Bhilai}
  \state{Chhattisgarh}
  \country{India}
}
\email{vikrantsahu@iitbhilai.ac.in}

\author{Dr. Gagan Raj Gupta}
\affiliation{%
  \institution{Indian Institute of Technology}
  \city{Bhilai}
  \state{Chhattisgarh}
  \country{India}
}
\email{gagan@iitbhilai.ac.in}

\author{Dr. Vandita Singh}
\affiliation{%
  \institution{All India Institute of Medical Sciences}
  \city{Rajkot}
  \state{Gujarat}
  \country{India}
}
\email{vandita300@gmail.com }


\renewcommand{\shortauthors}{Vissapragada et al.}

\begin{abstract}

For doctors to truly trust artificial intelligence, it can't be a black box. They need to understand its reasoning, almost as if they were consulting a colleague. We created HistoLens\footnote{The source code for HistoLens is available at: \url{https://github.com/Sandeep-4469/HistoLens}} to be that transparent, collaborative partner. It allows a pathologist to simply ask a question in plain English about a tissue slide—just as they would ask a trainee. Our system intelligently translates this question into a precise query for its AI engine, which then provides a clear, structured report. But it doesn't stop there. If a doctor ever asks, ``Why?'', HistoLens can instantly provide a `visual proof' for any finding—a heatmap that points to the exact cells and regions the AI used for its analysis. We've also ensured the AI focuses only on the patient's tissue, just like a trained pathologist would, by teaching it to ignore distracting background noise. The result is a workflow where the pathologist remains the expert in charge, using a trustworthy AI assistant to verify their insights and make faster, more confident diagnoses.

\end{abstract}



 %\includegraphics[width=\textwidth]{sampleteaser}
 %\caption{Seattle Mariners at Spring Training, 2010.}
 %\Description{Enjoying the baseball game from the third-base
 %seats. Ichiro Suzuki preparing to bat.}
 %\label{fig:teaser}


\maketitle
\section{Introduction}

The growing adoption of Vision-Language Models (VLMs) in clinical workflows promises to revolutionize histopathology by automating complex diagnostic tasks \cite{hartsock2024vlm, conch2024}. However, this powerful technology faces two critical barriers that prevent its widespread adoption. The first is a profound **trust gap**: most VLMs operate as \textit{“black-box”} systems, delivering a final report with little insight into their reasoning. This opacity is clinically untenable, as a pathologist cannot be expected to take professional responsibility for a diagnostic score without understanding the underlying visual evidence. The second is a **prompting gap**: these advanced models often require precisely formatted prompts, a technical hurdle that distances the clinical expert from the AI tool and hinders seamless integration into the diagnostic workflow.

The severity of this trust gap becomes clear when considering high-stakes clinical applications. In modern oncology, the quantitative analysis of immunohistochemical (IHC) markers is essential for patient care. For instance, the Ki-67 labeling index, a measure of cellular proliferation, is critical for tumor grading and prognosis. Similarly, scoring the expression of PD-L1, an immune checkpoint protein, is vital for guiding life-saving immunotherapy decisions \cite{binder2024inclmed}. An opaque AI providing a score for these markers without justification is clinically unacceptable, as even small variations in quantification can significantly alter a patient's treatment pathway. The need for a verifiable, trustworthy, and usable AI is therefore not just a technical challenge, but a clinical necessity.

To address this critical need, we present HistoLens, an intelligent framework designed to transform VLMs from opaque analytical engines into transparent, interactive partners. We bridge the trust and prompting gaps with a multi-faceted approach. The primary contributions of this work are as follows:
\begin{enumerate}
    \item \textbf{A Multi-Modal XAI Toolkit:} An interactive suite that allows clinicians to visually probe any VLM finding, providing a spectrum of explainability from high-level regional "hotspots" down to the fine-grained cellular features that influenced the model's decision.
    \item \textbf{A Novel Method for Mitigating Shortcut Learning:} We demonstrate how the XAI toolkit can be used to diagnose critical "shortcut learning" flaws \cite{geirhos2020shortcut} in the VLM, transforming it from a passive viewer into an active tool for AI model auditing and debugging. We introduce Region-of-Interest (ROI) In-painting as a robust technique to correct these flaws.
    \item \textbf{A Semantic Prompt Synthesizer:} A module powered by a local Llama 3 model that translates a clinician's natural-language query (e.g., "What is the Ki-67 index?") into the perfectly structured prompt required by the VLM, creating an intuitive conversational interface.
\end{enumerate}
Unlike prior XAI frameworks, HistoLens unifies prompt synthesis, shortcut mitigation, and visual explainability into a single interactive clinical workflow — enabling both transparency and control for end users. HistoLens is not merely a viewer but an essential diagnostic suite for the AI model itself, fostering the trust and collaboration necessary for the responsible integration of AI into real-world clinical practice.

\section{Related Work}
HistoLens lies at the intersection of Vision-Language Models for medicine, Explainable AI (XAI), diagnosing model flaws like shortcut learning, and emerging approaches to Human-AI collaboration.

\subsection{Vision-Language Models in Medicine}
Foundation models promise “generalist medical AI” \cite{moor2023foundation}. Vision-Language Models (VLMs), which jointly learn from images and text, are central to this effort. Architectures like LLaVA \cite{liu2023llava}, combining a vision encoder with a large language model, have shown success in medical VQA and image summarization \cite{hartsock2024vlm}. In pathology, systems like PathAlign \cite{ahmed2024pathalign} and CONCH \cite{conch2024} demonstrate VLMs’ utility on whole slide images. MedGemma \cite{medgemma2025}, used in our work, exemplifies this trend with strong zero-shot reasoning on medical imagery. However, most VLM research emphasizes performance while overlooking transparency and verifiability in clinical practice \cite{nauman2025pathvlm}. HistoLens directly addresses this gap by making reasoning interpretable and auditable.

\subsection{Explainable AI for Medical Vision}
The opacity of deep learning has driven extensive work in XAI. Heatmap-based methods like CAM \cite{zhou2016cam} and Grad-CAM \cite{selvaraju2017gradcam} remain standard. Extensions such as Grad-CAM++ \cite{chattopadhyay2018gradcamplus} and HiResCAM \cite{draelos2020hirescam} improve localization and resolution, while pixel-level methods like Guided Backpropagation \cite{springenberg2015guided} capture fine-grained cues. Rather than treating these approaches as interchangeable, HistoLens integrates them into a multi-modal toolkit, enabling users to move from regional to pixel-level explanations in one interactive workflow.

\subsection{Diagnosing and Mitigating Model Flaws}
AI models often exploit spurious correlations—shortcut learning—rather than true medical concepts \cite{geirhos2020shortcut, ye2024spurious}. This poses critical risks when models rely on artifacts like slide borders or scanner text \cite{boland2024shortcuts, macdonald2024shortcut}. While prior work documents these flaws, few tools let clinicians uncover and correct them in practice \cite{dhahri2024detecting}. HistoLens introduces ROI In-painting, a domain-specific intervention that replaces distracting background with a neutral fill, reducing shortcut signals. Related to masking and inpainting approaches in medical imaging \cite{jin2024maskmedpaint, wang2024lesion, silva2020inpainting}, our method is explicitly designed for interactive, expert-driven debugging.

\section{Approach}
\begin{figure}[h]
\small
  \centering
  \includegraphics[width=\linewidth]{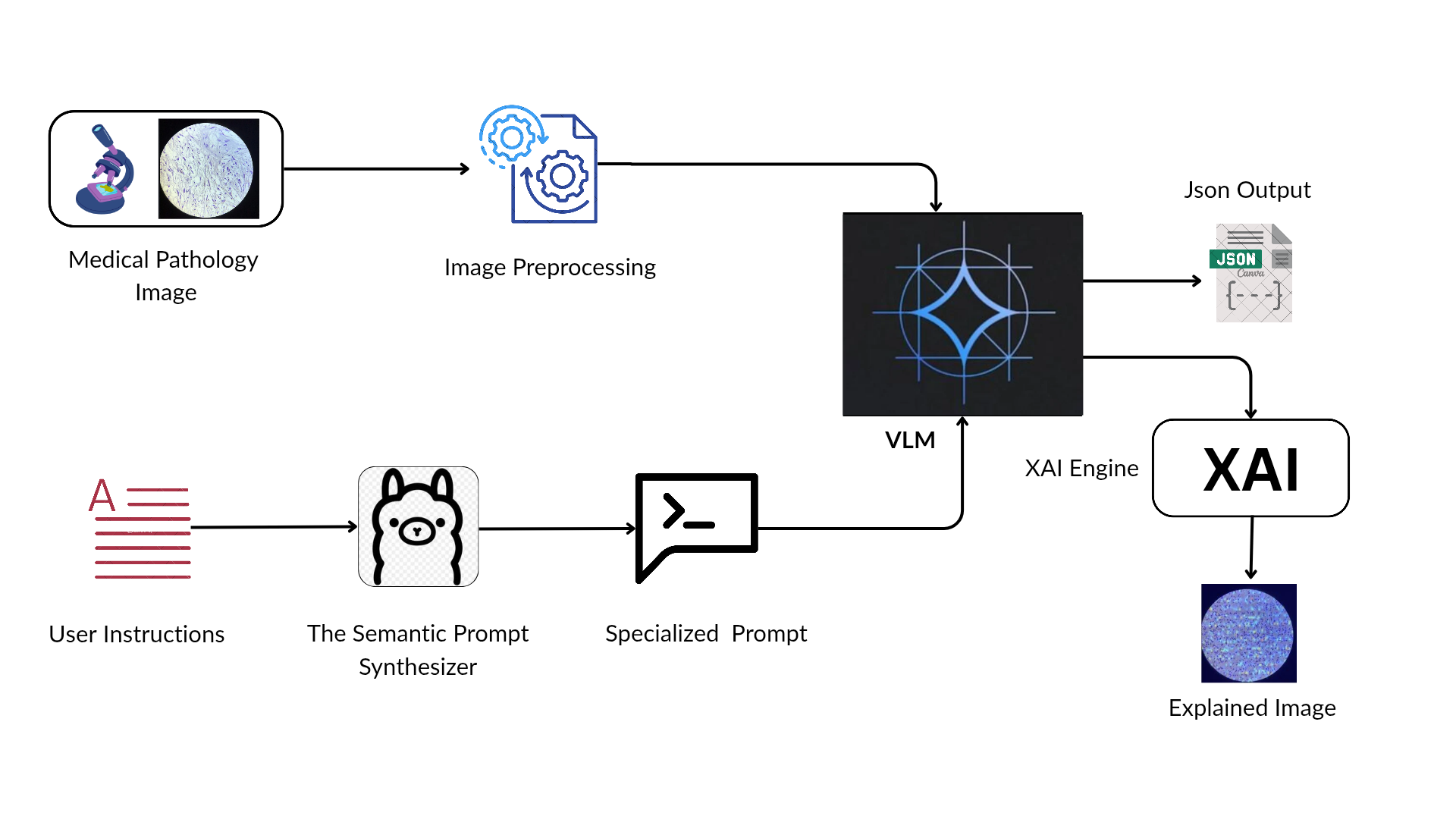}
  \caption{"The HistoLens workflow. A pathologist's natural language query about a Ki-67 stained slide is converted by the Semantic Prompt Synthesizer, analyzed by the VLM Core, and the result is visualized with the XAI Engine, allowing for full transparency and verification."}
  \Description{A flowchart showing the HistoLens architecture. It starts with a User Query, goes to the Semantic Prompt Synthesizer, then to the VLM Analysis Core (with an optional ROI Inpainting step), and finally to the Multi-Modal XAI Engine, which produces a report and visual explanations.}
  \label{fig:architecture}
\end{figure}

HistoLens system is architected as a modular, multi-stage pipeline designed to create a seamless workflow from a clinician's initial query to a fully verifiable, AI-generated analysis. The framework integrates three core pillars: a Semantic Prompt Synthesizer, a VLM Analysis Core, and a Multi-Modal XAI Engine, as depicted in Figure \ref{fig:architecture}. The entire system is designed to produce outputs that are not only computationally sound but also clinically relevant, with all VLM-generated reports benchmarked against evaluations from an expert pathologist at AIIMS.
\subsection{The Semantic Prompt Synthesizer}
To bridge the ``prompting gap'' for clinicians, HistoLens incorporates a Semantic Prompt Synthesizer. This module is powered by a locally-hosted Llama 3 8B model, served via Ollama to ensure data privacy. When a user enters a natural-language clinical query (e.g., ``are there many strongly stained immune cells?''), the query is embedded within a carefully engineered meta-prompt. This meta-prompt provides the LLM with its persona, strict output formatting rules, and a high-quality few-shot example of a successful query-to-prompt transformation. The \texttt{generate\_professional\_prompt} function sends this request to the Llama 3 model with a low temperature for deterministic output. The initialization prompt explicitly defines the model’s clinical persona (e.g., “You are a pathology assistant...”) and enforces a structured JSON schema to ensure consistent reasoning.  

\begin{table*}[ht]
\caption{Example of the Semantic Prompt Synthesizer in Action. The table illustrates how a pathologist's natural-language query is automatically transformed into a structured, domain-specific prompt for the VLM.}
\label{tab:prompt_synthesis}
\begin{tabularx}{\textwidth}{|l|X|}
\hline
\textbf{Component} & \textbf{Content} \\
\hline
\hline
\textbf{User's Prompt} & this is pdl-1 stain image and belongs to brain tissue. give me complete details \\
\hline
\textbf{Generated Specialized Prompt} & \textbf{System Prompt:} You are a pathology assistant specialized in analyzing stained histopathology images, including PDL1 immunohistochemistry. Please analyze the provided image of brain tissue and return your findings in the following JSON format. \newline\newline \textbf{Notes:} Tumor cells may appear lightly stained while normal brain parenchymal cells may appear heavily stained. Ensure accurate distinction. Be careful to exclude non-relevant glial cells if present. \newline\newline \textbf{Required JSON Structure:} \newline \texttt{\{"stain\_type": "PDL1", "percentage\_of\_cells\_stained": "0-100", ... \}} \\
\hline
\textbf{Final VLM Output (JSON)} & 
\texttt{\{ \newline}
\texttt{  "stain\_type": "PDL1", \newline}
\texttt{  "percentage\_of\_cells\_stained": "0-10", \newline}
\texttt{  "type\_of\_cells\_stained": "tumor cells", \newline}
\texttt{  "staining\_location\_per\_cell": "cytoplasmic", \newline}
\texttt{  "report": "PDL1 immunohistochemistry shows a low percentage of tumor cells exhibiting cytoplasmic staining.", \newline}
\texttt{  "explanation": "The image shows a tissue sample with a predominantly cellular appearance... the low PDL1 expression suggests a less aggressive tumor." \newline}
\texttt{\}} \\
\hline
\end{tabularx}
\end{table*}

\subsection{The VLM Analysis Core}
At the heart of HistoLens lies the MedGemma-4B-IT model \cite{medgemma2025}, which we chose after experimenting with several recent vision-language frameworks. MedGemma showed strong zero-shot reasoning on diagnostic imagery and, importantly, has been pre-trained on a wide range of medical data. This domain familiarity allows it to interpret stain-specific visual patterns more reliably than general-purpose VLMs. Its medium scale (around 4B parameters) offered a practical balance between interpretability, visual precision, and compute efficiency — an aspect that matters in real-world hospital systems. To ensure reproducibility, all inference runs used deterministic greedy decoding \texttt{(do\_sample=False)}.  

Before analysis, users can optionally enable our ROI In-painting pre-processing technique. This step is our direct intervention to reduce the ``shortcut learning'' artifacts we observed during early testing. The \texttt{apply\_roi\_inpainting} function detects the main tissue sample, computes its average color, and replaces irrelevant background with a uniform fill. In practice, this encourages the model to focus on genuine pathological structures rather than surrounding noise. The final analysis is then performed on this cleaned image, producing more stable and clinically reliable results.

\subsection{The Multi-Modal XAI Engine}
To close the "interaction gap" and make the VLM's reasoning transparent, our XAI Engine provides visual evidence for any claim made in the VLM's report. The technical implementation is designed for robustness and precision.
\subsubsection{Targeted Loss and Unified Gradient Context}
When a user selects a specific finding from the JSON report (e.g., \texttt{"staining-\\\_intensity-\_grade": 3}), we calculate a loss based only on the corresponding token sequence. This ensures the explanation is sharply focused on the evidence for that specific claim. To guarantee correct gradient capture, our \texttt{generate\_explanation} function temporarily switches the model to \texttt{train()} mode, performs a single, unified backward pass, and uses a \texttt{finally} block to always return the model to \texttt{eval()} mode.
\begin{figure}[htbp] % Use [htbp] for better placement flexibility
  \centering
  \includegraphics[width=\linewidth]{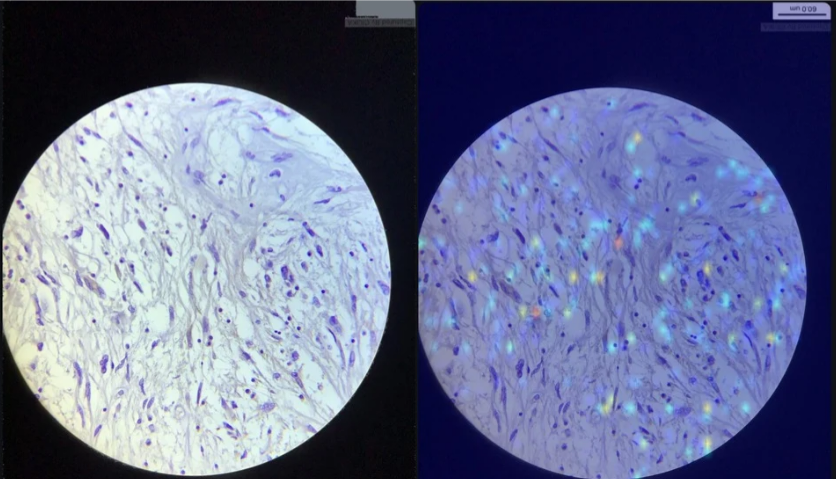} % <-- Replace with your merged image file
  \caption{Visual verification using the HistoLens XAI toolkit. The left panel shows the original PD-L1 stained input image. The VLM identified the \texttt{"staining\_location\_per\_cell"} as cytoplasmic. The right panel shows the corresponding Grad-CAM heatmap, which confirms the model correctly focused on the cytoplasm of the tumor cells (highlighted in red/yellow), increasing the pathologist's trust in the output.}
  \label{fig:gradcam_example}
  \Description{A side-by-side comparison of a histopathology slide and its AI-generated explanation heatmap.}
\end{figure}

\subsubsection{The Toolkit}
All our CAM-based methods target the final layer of the vision encoder (\texttt{model.vision\_tower.vision\_model-\\.encoder.layers[-1]}) to capture high-level semantic features. The toolkit provides a suite of complementary views:
\begin{itemize}
    \item \textbf{Grad-CAM \cite{selvaraju2017gradcam}:} For a high-level overview of important regions.
    \item \textbf{Grad-CAM++ \cite{chattopadhyay2018gradcamplus}:} For better localization of multiple, scattered objects.
    \item \textbf{HiResCAM \cite{draelos2020hirescam}:} For cleaner heatmaps with sharper boundaries.
    \item \textbf{Guided Grad-CAM:} Fuses the regional context of Grad-CAM with a pixel-precise saliency map from Guided Backpropagation \cite{springenberg2015guided}, allowing for an unparalleled deep dive into the specific textures and cell boundaries that influenced the VLM's decision.
\end{itemize}

\section{Dataset and Validation}
To rigorously evaluate the HistoLens framework, we curated a representative dataset of 60 histopathology images, designed to mirror the diversity and complexity of real-world clinical samples.

\subsection{Dataset Composition}
The dataset comprises three cohorts of 20 images each, corresponding to three of the most clinically significant immunohistochemical (IHC) stains used in modern oncology:
\begin{itemize}
    \item \textbf{Ki-67:} A critical marker for assessing tumor cell proliferation.
    \item \textbf{BRAF:} A key biomarker for targeted therapy in melanoma and other cancers.
    \item \textbf{PD-L1:} An essential predictive marker for guiding immunotherapy decisions.
\end{itemize}

All images were collected in JPEG (.jpg) format to ensure broad compatibility. The dataset was intentionally designed to include both inter-stain variability (reflecting different biomarker targets and protocols) and intra-stain variability (e.g., differences in staining intensity, tissue morphology, and background artifacts). This diversity ensures that our evaluation robustly tests the system's performance under realistic conditions.

\subsection{Expert Annotation and Clinical Validation}
To establish a reliable ground truth, all images in the dataset were independently reviewed and annotated by expert pathologists, ensuring staining quality, accurate identification of diagnostically relevant regions, and correct interpretation of biomarker expression. All patient identifiers were fully anonymized, and the dataset was organized into stain-specific folders for reproducibility.  

Beyond dataset preparation, we conducted a formal clinical validation of HistoLens by comparing its structured JSON outputs (e.g., \texttt{staining\_intensity\_grade}, \texttt{type\_of\_cells\_stained}) against expert assessments from a senior pathologist at the All India Institute of Medical Sciences (AIIMS). The evaluation focused on two axes: (i) \textit{Clinical Accuracy}—whether the VLM’s analysis aligned with expert readings, and (ii) \textit{Report Quality}—whether the narrative outputs were coherent, clinically relevant, and free of hallucinations.  

Quantitatively, HistoLens achieved an \textbf{86.7\% agreement rate} with expert annotations and demonstrated a \textbf{21\% improvement in focus consistency} when ROI In-painting was enabled. Importantly, no signs of overfitting were observed, as the model’s attention patterns and reasoning remained stable across different stain categories. Interestingly, in a subset of PD-L1 slides, the model occasionally confused nuclear and cytoplasmic staining patterns—a subtle distinction that even experienced pathologists find challenging due to morphological overlap. These borderline cases reflect the inherent ambiguity of immunohistochemical interpretation rather than a model-specific error.  

This dual role of expert annotation and validation both grounds our experiments in trustworthy clinical labels and substantiates our thesis that HistoLens can diagnose and mitigate reasoning flaws, producing outputs that are demonstrably more reliable for clinical use. Figure~\ref{fig:gradcam_example} shows a representative dataset image.

% \section{Future Work}
% The HistoLens system opens several promising avenues for future research and development focused on enhancing interactivity, broadening the system's applicability, and performing rigorous clinical validation.

% \subsection{Human-in-the-Loop Correction}
% Enabling real-time, human-in-the-loop correction so that a pathologist can directly interact with the XAI heatmaps to perform a rapid, on-the-fly fine-tuning of the model's projector module, allowing the AI to learn from expert guidance within a single session.

% \subsection{Formal Clinical User Study}
% Conducting a formal user study with a cohort of pathologists to quantify the impact of HistoLens on clinical practice aimed to measure KPIs like diagnostic accuracy, time-to-diagnosis, and user confidence when using HistoLens while providing crucial qualitative feedback for refining the XAI toolkit.

% \subsection{Cross-Modal and Cross-Domain Generalization}
% Integrating other SOTA medical VLMs to test the framework's generalizability and adapting the HistoLens to other medical imaging domains, such as radiology and dermatology demonstrating its potential as a universal platform for building trust in medical AI across different specializations.

\section{Demonstration}
\noindent\textbf{A video demonstration of the HistoLens workflow is available at:}  
\textbf{\url{https://youtu.be/szO414pjHsI}}

The demo showcases the following steps:  
\begin{itemize}
    \item \textbf{Human Query:} The pathologist enters a natural language prompt.
    \item \textbf{Prompt Refinement:} The Semantic Prompt Synthesizer (LLaMA) converts it into a precise, professional query.
    \item \textbf{AI Analysis:} The MedGemma-4B model processes the query and outputs a structured JSON report containing:
    \begin{itemize}
        \item Stain type
        \item Percentage of cells stained
        \item Stain grade
        \item Findings and explanation
        \item Stain locations
    \end{itemize}
    \item \textbf{Explainability:} The pathologist selects any key from the JSON and requests an explanation.
    \item \textbf{Heatmap Generation:} By choosing Grad-CAM, Grad-CAM++, HiResCAM, or Guided Grad-CAM, HistoLens produces a corresponding heatmap highlighting the exact regions used for analysis.
\end{itemize}

\section{Conclusion and Future Directions}
HistoLens tackles one of the most persistent challenges in clinical AI — the question of trust. By addressing the “prompting gap’’ through a Semantic Prompt Synthesizer and the “interaction gap’’ through a multimodal explainability toolkit, it transforms opaque models into interpretable, verifiable systems. Rather than functioning only as a visualization layer, HistoLens behaves as a diagnostic companion that can reveal and even correct reasoning flaws in advanced VLMs through ROI In-painting.  

Although our current dataset includes 60 carefully curated and annotated slides, the results offer a convincing proof of concept for transparent, clinically aligned reasoning. In future iterations, we plan to extend validation across multiple institutions, explore other VLM–LLM pairings such as CONCH and PathAlign, and conduct formal user studies to measure the impact of HistoLens on diagnostic efficiency and clinician confidence.  

Ultimately, we view HistoLens not just as a histopathology tool but as a foundation for trustworthy human–AI collaboration in medicine.

\bibliographystyle{ACM-Reference-Format}
\bibliography{sample-base}

\end{document}